\documentclass[12pt]{article}
\usepackage{arxiv}
\usepackage[a4paper,top=3cm,bottom=2cm,left=2cm,right=2cm,marginparwidth=1.25cm]{geometry}

\usepackage{tikz}
\usepackage[utf8]{inputenc}
\usepackage{pgfplots}
\DeclareUnicodeCharacter{2212}{−}
\usepgfplotslibrary{groupplots,dateplot}
\usetikzlibrary{patterns,shapes.arrows, positioning,matrix}
\pgfplotsset{compat=newest}

\usepackage{microtype}
\usepackage{graphicx}
\usepackage{subfigure}
\usepackage{booktabs}
\usepackage{hyperref}
\usepackage{amsmath}
\usepackage{amssymb}
\usepackage{mathtools}
\usepackage{amsthm}
\usepackage[capitalize,noabbrev]{cleveref}

\usepackage{times}

\usepackage{xcolor} 
\usepackage{algorithm}
\usepackage{algorithmic}

\usepackage[numbers]{natbib}
\usepackage{eso-pic} 
\usepackage{forloop}
\usepackage{url}

\usepackage{caption}
\newcommand{\yrcite}[1]{\cite{#1}}

\newcommand{\legendsize}{10cm}
\newcommand{\legendsizelarge}{14cm}
\newcommand{\diagramninetydegree}{-5.02}
\newcommand{\diagramlgantag}{4.87em}
\newcommand{\diagramlgan}{3.04em}

\begin{document}

\title{Imbalanced Classification via a Tabular Translation GAN}

\date{} 					

\author{ \hspace{1mm}Jonathan~Gradstein \textsuperscript{1}
	\And
	\hspace{1mm}Moshe~Salhov \textsuperscript{1,2}\\
    \And
	\hspace{1mm}Yoav~Tulpan \textsuperscript{2}\\
	\And
	\hspace{1mm}Ofir~Lindenbaum \textsuperscript{3}\\
	\And
	\hspace{1mm}Amir~Averbuch \textsuperscript{1}\\
}


\renewcommand{\shorttitle}{Imbalanced Classification via a Tabular Translation GAN}

\ifdefined\nohyperref\else\ifdefined\hypersetup
\definecolor{mydarkblue}{rgb}{0,0.08,0.45}
\hypersetup{ %
pdftitle={Imbalanced classification via a tabular translation GAN},
linkcolor=mydarkblue,
    citecolor=mydarkblue,
    filecolor=mydarkblue,
    urlcolor=mydarkblue
}

\maketitle

\begin{abstract}
When presented with a binary classification problem where the data exhibits severe class imbalance, most standard predictive methods may fail to accurately model the minority class. We present a model based on Generative Adversarial Networks which uses additional regularization losses to map majority samples to corresponding synthetic minority samples. This translation mechanism encourages the synthesized samples to be close to the class boundary. Furthermore, we explore a selection criterion to retain the most useful of the synthesized samples. Experimental results using several downstream classifiers on a variety of tabular class-imbalanced datasets show that the proposed method improves average precision when compared to alternative re-weighting and oversampling techniques. 
\end{abstract}


\section{Introduction}
Data that exhibits class imbalance appears frequently in real-world scenarios \cite{bauder2018empirical}, in varying domains and applications: detecting pathologies or diseases in medical records \cite{zhao2018framework}, preventing network attacks in cybersecurity \cite{wheelus2018tackling}, detecting fraudulent financial transactions \cite{makki2019experimental}, distinguishing between earthquakes and explosions \cite{rabin2016earthquake} and detecting spam communications \cite{tang2006fast}. In addition to these applications, where the class distribution is naturally skewed due to the frequency of events, some applications may exhibit class imbalance caused by extrinsic factors such as collection and storage limitations \cite{he2009learning}.

Most standard classification models are designed around and implicitly assume a relatively balanced class distribution; when applied without proper adjustments they may fail to accurately model the minority class and converge on a solution that over-classifies the majority class due to its increased prior probability \cite{johnson2019survey}. These models thus neglect recall on the minority class and lead to unsatisfactory results when we desire high performance on a more balanced testing criterion. This issue is exacerbated by the fact that commonly used metrics such as accuracy may be misleading in evaluating the performance of the model. Even models that naively classify all samples as majority may have high accuracy under severe class imbalance.
Most approaches to dealing with these shortcomings fall broadly into two categories: re-weighting the loss objective to more heavily account for the minority class, and resampling the input dataset such that the minority class is more prominent. 

The proposed approach falls in the latter category, and provides a technique to oversample the minority class by generating synthetic minority samples that correspond to translations of real samples from the majority class. Leveraging the diversity of the real majority samples allows this translation approach to generate samples that are beneficial to classification performance.
We focus our attention on the problem of class-imbalanced binary classification for tabular datasets; many of the aforementioned real-world scenarios are instances of this problem.

\subsection{Contribution}
In this work, our contribution is to achieve the following goals:

\begin{itemize}
	\item Propose a novel GAN-based model to generate synthetic minority samples.
	\item Leverage the diversity of the majority class by ensuring that the generated points are translated from the majority class.
	\item Provide a strategy for selection of the most useful subset of these translated synthetic samples.
	\item Verify that this process compares favorably to alternative class-imbalanced techniques in terms of downstream balanced classification performance on tabular datasets.
\end{itemize}

\subsection{Related work}
\paragraph{Background}
Methods to address class imbalance have been extensively studied. A straightforward approach involves re-weighting the cost/penalty associated with the classes in proportion to their cardinality; in all the models and experiments below, we apply this loss re-weighting technique implicitly when the classes are imbalanced. Alternatively or in addition, resampling the input training data distribution is a common technique. The simplest forms of this technique are random oversampling, which duplicates samples from the minority, and random undersampling, which discards samples from the majority. Undersampling risks losing valuable information in the process, while oversampling introduces the risk of overfitting \cite{chawla2004special}.

\paragraph{Oversampling}
Methods more sophisticated than random oversampling have been introduced in the last few decades, notably SMOTE \cite{chawla2002smote} and its variants, which generate new minority samples by interpolating nearby training samples. One such variant, Borderline SMOTE \cite{han2005borderline} is similar to the proposed method in the sense that it oversamples specifically near the class boundary region. SUGAR \cite{lindenbaum2018geometry} takes a geometric approach, generating new samples along a manifold learned through a diffusion process.

\paragraph{Image-to-image translation}
Unpaired, unsupervised translation techniques have become prevalent in the deep learning literature in recent years particularly for the visual modality \cite{zhu2017unpaired, kim2017learning, choi2018stargan}. Such techniques usually rely on a cyclic loss to maintain translation correspondence in absence of supervision. The model presented by Choi et al. \yrcite{choi2018stargan} is also able to perform the translation across more than two domains. These works do not explore the prospect of imbalanced classification between the domains, and do not utilize the synthesized samples to train a further classifier. Some approaches exist that aim to augment a minority class, but they are mostly focused on visual image tasks \cite{lee2020generative, zhang2021deep}. 


\paragraph{Oversampling via translation} Translation-based approaches to oversampling are relatively underdeveloped. The most comparable approach is M2m \cite{kim2020m2m}, where samples are translated by traversing along the path of the gradient of an initial baseline classifier. As the authors demonstrate, adding these translated samples improves classification performance on the tested datasets. In practice, the authors find that many of the synthetic samples are adversarial samples: they are close to the input sample in the high-dimensional feature space. During our experimentation, this approach yields unsatisfactory results in generating informative minority samples on \textit{tabular} datasets, including toy synthetic datasets. We hypothesize that this is due to the feature heterogeneity and low dimensionality of tabular datasets. We then attempted to modify the approach by making more flexible gradient steps: instead of using a constant stepsize as used in the original paper, we adjusted the gradient stepsize by using optimization methods \cite{kingma2014adam}. This did not yield significantly better results (see \cref{fig:m2m} for an example).

\begin{figure}
	\centering
    \input{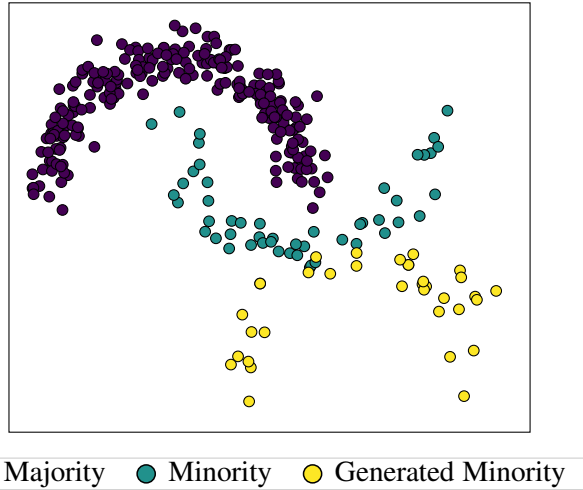}
	\caption{An M2m-style approach does not generate desirable samples on low-dimensional synthetic datasets.}
	\label{fig:m2m}
\end{figure}

\newsavebox\initialmajbox
\savebox\initialmajbox{%
\usetikzlibrary{pgfplots.groupplots}
\pgfplotsset{colormap/viridis}
\definecolor{majcol}{RGB}{68,1,84}
\definecolor{mincol}{RGB}{33,145,140}
\definecolor{gencol}{RGB}{253,231,37}

\begin{tikzpicture}[yscale=1, xscale=1]

\begin{axis}[
ticks=none,
xmin=-1, xmax=0.6,
ymin=-1.15, ymax=2.2,
  axis lines=none,
  width=3cm,
  height=6cm
]
\addplot[only marks, scatter, scatter src=explicit symbolic, scatter/classes={
            0.0={color of colormap={0}, draw=black},
            1.0={color of colormap={500}, draw=black},
            2.0={color of colormap={1000}, draw=black}
        }]
table [x=x, y=y, meta=colordata]{%
x  y  colordata
-0.405972628032873 1.08943142367152 0.0
-0.209441861168637 1.0040342706309 0.0
-0.279224444627808 0.982969428888976 0.0
-0.138397109307423 0.98875499962785 0.0
-0.284533995497052 0.985184433487537 0.0
-0.366387894168781 0.914604129136502 0.0
-0.447445261337915 1.01573524215079 0.0
-0.320134963883679 0.964006586436263 0.0
-0.32578807505877 0.922552134346806 0.0
-0.334310689999376 0.985751869981636 0.0
-0.334813235851992 0.862713528377198 0.0
-0.539992065103079 0.980147811419253 0.0
-0.510097837771451 0.837093355752808 0.0
-0.617316905922986 0.900582522747724 0.0
-0.506460776275677 0.810541544575402 0.0
-0.599331947221634 0.864556845863326 0.0
-0.474877672854972 0.742110398822968 0.0
-0.325914080502624 0.739869003456383 0.0
-0.600596644162408 0.708375019073441 0.0
-0.503677113644801 0.784077708317368 0.0
-0.55050608263537 0.618877802277663 0.0
-0.518103336544946 0.6532848903867 0.0
-0.715762741993136 0.61126115550801 0.0
-0.665411190540856 0.566381031122409 0.0
-0.567813817362793 0.465679111846018 0.0
-0.658320105104442 0.595853944256992 0.0
-0.782523745517987 0.566844854100172 0.0
-0.62153405895218 0.551244658028133 0.0
-0.547912227588558 0.483837168568771 0.0
-0.594065999335881 0.375381065612633 0.0
-0.837307814801221 0.421271300136321 0.0
-0.958647210165837 0.321748544594155 0.0
-0.642147324299436 0.260113898822335 0.0
-0.633198080693065 0.263263917578599 0.0
-0.774754623136307 0.238876138670322 0.0
-0.605551118292851 0.126751351947146 0.0
-0.764361696773754 0.131010189394104 0.0
-0.704551213191154 0.123443216288088 0.0
-0.845863400649068 0.0917732891996059 0.0
-0.809590742812206 -0.0499059874669636 0.0
-0.640082916067684 -0.0884156382926711 0.0
-0.755715763676115 -0.0289112707821153 0.0
-0.77920488261779 -0.0920006382028853 0.0
-0.569996284572094 -0.13994075549263 0.0
-0.648674685734265 -0.323784783267186 0.0
-0.668122710343483 -0.246532745030612 0.0
-0.686712702102506 -0.303616441451836 0.0
-0.724681550160926 -0.322585695362556 0.0
-0.510315458859828 -0.297665698560793 0.0
-0.802567031088085 -0.383874228249985 0.0
-0.718353832222415 -0.361811034682468 0.0
-0.790803544295897 -0.445532545125191 0.0
-0.780050977640391 -0.415193949205849 0.0
-0.727088807254081 -0.537964586321177 0.0
-0.796373456665784 -0.512941672913472 0.0
-0.564659661447767 -0.603924396464756 0.0
-0.745842168172048 -0.630885046767032 0.0
-0.518485343439726 -0.675591671457914 0.0
-0.599780214354183 -0.628522480834802 0.0
-0.530715019124051 -0.741754067103015 0.0
-0.69278571468368 -0.761789444336552 0.0
-0.59093769856132 -0.792231359838104 0.0
-0.562399695128152 -0.768769794872921 0.0
-0.694002316703443 -0.808188613298762 0.0
-0.443399409014069 -0.859724576958381 0.0
-0.620969894592015 -0.85738786702246 0.0
-0.482595934310669 -0.828579547377016 0.0
-0.375915766044089 -0.860346952506761 0.0
-0.537619769463504 -0.928703433022439 0.0
-0.505277240146128 -0.948229086985989 0.0
-0.352706368496406 -0.981958233297444 0.0
-0.535929512857102 -0.943419740671978 0.0
-0.412379158274071 -0.997306194203291 0.0
-0.493443578402708 -0.950371002547921 0.0
-0.409653769804706 -0.938505171695537 0.0
-0.433499438682477 -1.03049076628318 0.0
-0.360727113600148 -1.05393716969678 0.0
-0.329638372533288 -0.996716630915234 0.0
-0.163448825572807 -1.03548107124723 0.0
-0.404229136435543 -1.0070310013508 0.0
-0.287784411964846 -0.0397957346611695 1.0
-0.0395201915728942 -0.0045068856074113 1.0
0.00161506831413982 0.167879797093616 1.0
0.148940664228943 0.00178756394267521 1.0
0.152260616587067 0.0804453212579361 1.0
0.274285381989209 0.249788264456903 1.0
0.307483193012641 0.271348716044172 1.0
0.454160113460011 0.384770520503727 1.0
0.379940927053089 0.438218168794694 1.0
0.40853186291228 0.669264173502633 1.0
0.493392738672171 0.827651826357994 1.0
0.384662408567469 0.843917358152303 1.0
0.431939902345132 0.898480322311085 1.0
0.303119265801782 1.07950822827985 1.0
0.453821643554384 1.29121053497301 1.0
0.512006316742393 1.35325835138112 1.0
0.398108836984787 1.49361189710571 1.0
0.481133255015837 1.63105047694138 1.0
0.140833594344003 1.80653126517708 1.0
0.324196318630291 1.7885914603562 1.0
0.231321970943864 1.84994616912983 1.0
0.0708573625053293 1.92294814673344 1.0
-0.174481841747028 2.00437408417365 1.0
-0.0839265469398911 2.04880810102944 1.0
-0.0538597554466167 2.01941252071424 1.0
};

\end{axis}

\end{tikzpicture}
\centering
}

\newsavebox\generatedbox
\savebox\generatedbox{%
\usetikzlibrary{pgfplots.groupplots}
\pgfplotsset{colormap/viridis}
\definecolor{majcol}{RGB}{68,1,84}
\definecolor{mincol}{RGB}{33,145,140}
\definecolor{gencol}{RGB}{253,231,37}

\begin{tikzpicture}[yscale=1, xscale=1]

\begin{axis}[
ticks=none,
xmin=-1, xmax=0.6,
ymin=-1.15, ymax=2.2,
  axis lines=none,
  width=3cm,
  height=6cm
]
\addplot[only marks, scatter, scatter src=explicit symbolic, scatter/classes={
            0.0={color of colormap={0}, draw=black},
            1.0={color of colormap={500}, draw=black},
            2.0={color of colormap={1000}, draw=black}
        }]
table [x=x, y=y, meta=colordata]{%
x  y  colordata
-0.206109918896302 -0.118792439287932 2.0
-0.263796617806589 -0.0440271972638062 2.0
-0.180008822717377 -0.0191154820674229 2.0
-0.252714369128004 0.154753038875566 2.0
-0.146360861592727 0.04615033225229 2.0
-0.140314508794594 0.0136643392681508 2.0
-0.202985668285536 -1.81370036382855e-05 2.0
-0.221307773970668 0.0614601993754487 2.0
-0.0936331357267495 -0.0818846571689675 2.0
-0.0865610429787562 0.115324986870352 2.0
-0.162132944300565 0.198460036899865 2.0
-0.0292526426406232 0.0361984812934024 2.0
-0.00777980299574255 0.208245403890785 2.0
0.0804654347598988 0.0601761889381183 2.0
0.0519115942578465 0.155189807850933 2.0
-0.146499202132317 0.183015325353839 2.0
-0.0513686700205802 0.276312209647027 2.0
-0.152413788853327 0.352462640824186 2.0
-0.0247465423312647 0.225996413868418 2.0
-0.0233823287262425 0.380664489058929 2.0
0.108643970233493 0.445184079546015 2.0
0.035907044285989 0.310616618515652 2.0
0.150722271938072 0.238994398997356 2.0
0.12187681975218 0.346149495809248 2.0
0.00302710581971399 0.608693573497975 2.0
-0.00799972907374968 0.487757495383411 2.0
0.0877578527524834 0.445435398811882 2.0
0.01828696838057 0.548698958778968 2.0
0.0373057300995732 0.525902753983641 2.0
0.166296966643227 0.612088702569156 2.0
0.206190789080826 0.647154271234857 2.0
0.203345543638937 0.650537106521466 2.0
0.0683189756318581 0.803130091264154 2.0
0.200334091975137 0.82093063745688 2.0
0.132885945658044 0.826945886320008 2.0
0.0258144904661712 0.95708927835021 2.0
0.142307284518693 0.938282963261506 2.0
-0.0187149652218148 1.04389758349345 2.0
0.206325060112431 0.873771805978169 2.0
0.0801308162605947 1.00109140649329 2.0
0.129056682668186 1.19589150774701 2.0
0.234819081658577 0.996525924335586 2.0
0.212423027076362 1.06142827935846 2.0
0.117410084668763 1.2440327581476 2.0
0.0691623858912138 1.28299347539703 2.0
0.163742421947584 1.19492305082507 2.0
0.0246799892211049 1.26869152644438 2.0
0.210711185649519 1.24988421558422 2.0
-0.000887730676151222 1.44156097266813 2.0
0.0309595591197476 1.34938914351217 2.0
0.147936616729493 1.3668889877394 2.0
0.158659985194186 1.35612897011904 2.0
0.276662897515644 1.41363218572035 2.0
0.246472472734942 1.46984880777302 2.0
0.241482590086136 1.46937833328243 2.0
0.0133619094022552 1.73498956654963 2.0
0.073661509612148 1.61085135635185 2.0
0.0896570733437889 1.66791367917458 2.0
-0.0155228509651628 1.68476377591718 2.0
-0.0850596548402829 1.80031740231858 2.0
0.167024258943589 1.72230306340769 2.0
-0.00195782817901113 1.70226191537203 2.0
0.110580698810188 1.76140142738798 2.0
0.11993604776784 1.70398908604059 2.0
0.00749799112985539 1.88195945918958 2.0
0.1033485762785 1.84598835747455 2.0
-0.0987088259993369 1.90437115788428 2.0
-0.0166949295473942 2.00638625713888 2.0
0.119362542051087 1.8999028718953 2.0
0.0357118427422101 1.83917622779721 2.0
-0.145827473007981 1.93809032977626 2.0
-0.0925294079599508 1.91175394218614 2.0
-0.123300973475327 1.97559567313436 2.0
0.0291880662007094 1.78593436176704 2.0
-0.121518220146199 1.93231298007913 2.0
-0.185794389309373 1.95324958299531 2.0
-0.114203865166274 1.9172259501703 2.0
-0.148163235742273 1.99713853841754 2.0
-0.227345253647866 2.09377659211818 2.0
-0.114437347566115 2.00143396987072 2.0
};

\end{axis}

\end{tikzpicture}
\centering
}

\newsavebox\outputbox
\savebox\outputbox{%
\usetikzlibrary{pgfplots.groupplots}
\pgfplotsset{colormap/viridis}
\definecolor{majcol}{RGB}{68,1,84}
\definecolor{mincol}{RGB}{33,145,140}
\definecolor{gencol}{RGB}{253,231,37}

\begin{tikzpicture}[yscale=1, xscale=1]

\begin{axis}[
ticks=none,
xmin=-1, xmax=0.6,
ymin=-1.15, ymax=2.2,
  axis lines=none,
  width=3cm,
  height=6cm
]
\addplot[only marks, scatter, scatter src=explicit symbolic, scatter/classes={
            0.0={color of colormap={0}, draw=black},
            1.0={color of colormap={500}, draw=black},
            2.0={color of colormap={1000}, draw=black}
        }]
table [x=x, y=y, meta=colordata]{%
x  y  colordata
-0.405972628032873 1.08943142367152 0.0
-0.209441861168637 1.0040342706309 0.0
-0.279224444627808 0.982969428888976 0.0
-0.138397109307423 0.98875499962785 0.0
-0.284533995497052 0.985184433487537 0.0
-0.366387894168781 0.914604129136502 0.0
-0.447445261337915 1.01573524215079 0.0
-0.320134963883679 0.964006586436263 0.0
-0.32578807505877 0.922552134346806 0.0
-0.334310689999376 0.985751869981636 0.0
-0.334813235851992 0.862713528377198 0.0
-0.539992065103079 0.980147811419253 0.0
-0.510097837771451 0.837093355752808 0.0
-0.617316905922986 0.900582522747724 0.0
-0.506460776275677 0.810541544575402 0.0
-0.599331947221634 0.864556845863326 0.0
-0.474877672854972 0.742110398822968 0.0
-0.325914080502624 0.739869003456383 0.0
-0.600596644162408 0.708375019073441 0.0
-0.503677113644801 0.784077708317368 0.0
-0.55050608263537 0.618877802277663 0.0
-0.518103336544946 0.6532848903867 0.0
-0.715762741993136 0.61126115550801 0.0
-0.665411190540856 0.566381031122409 0.0
-0.567813817362793 0.465679111846018 0.0
-0.658320105104442 0.595853944256992 0.0
-0.782523745517987 0.566844854100172 0.0
-0.62153405895218 0.551244658028133 0.0
-0.547912227588558 0.483837168568771 0.0
-0.594065999335881 0.375381065612633 0.0
-0.837307814801221 0.421271300136321 0.0
-0.958647210165837 0.321748544594155 0.0
-0.642147324299436 0.260113898822335 0.0
-0.633198080693065 0.263263917578599 0.0
-0.774754623136307 0.238876138670322 0.0
-0.605551118292851 0.126751351947146 0.0
-0.764361696773754 0.131010189394104 0.0
-0.704551213191154 0.123443216288088 0.0
-0.845863400649068 0.0917732891996059 0.0
-0.809590742812206 -0.0499059874669636 0.0
-0.640082916067684 -0.0884156382926711 0.0
-0.755715763676115 -0.0289112707821153 0.0
-0.77920488261779 -0.0920006382028853 0.0
-0.569996284572094 -0.13994075549263 0.0
-0.648674685734265 -0.323784783267186 0.0
-0.668122710343483 -0.246532745030612 0.0
-0.686712702102506 -0.303616441451836 0.0
-0.724681550160926 -0.322585695362556 0.0
-0.510315458859828 -0.297665698560793 0.0
-0.802567031088085 -0.383874228249985 0.0
-0.718353832222415 -0.361811034682468 0.0
-0.790803544295897 -0.445532545125191 0.0
-0.780050977640391 -0.415193949205849 0.0
-0.727088807254081 -0.537964586321177 0.0
-0.796373456665784 -0.512941672913472 0.0
-0.564659661447767 -0.603924396464756 0.0
-0.745842168172048 -0.630885046767032 0.0
-0.518485343439726 -0.675591671457914 0.0
-0.599780214354183 -0.628522480834802 0.0
-0.530715019124051 -0.741754067103015 0.0
-0.69278571468368 -0.761789444336552 0.0
-0.59093769856132 -0.792231359838104 0.0
-0.562399695128152 -0.768769794872921 0.0
-0.694002316703443 -0.808188613298762 0.0
-0.443399409014069 -0.859724576958381 0.0
-0.620969894592015 -0.85738786702246 0.0
-0.482595934310669 -0.828579547377016 0.0
-0.375915766044089 -0.860346952506761 0.0
-0.537619769463504 -0.928703433022439 0.0
-0.505277240146128 -0.948229086985989 0.0
-0.352706368496406 -0.981958233297444 0.0
-0.535929512857102 -0.943419740671978 0.0
-0.412379158274071 -0.997306194203291 0.0
-0.493443578402708 -0.950371002547921 0.0
-0.409653769804706 -0.938505171695537 0.0
-0.433499438682477 -1.03049076628318 0.0
-0.360727113600148 -1.05393716969678 0.0
-0.329638372533288 -0.996716630915234 0.0
-0.163448825572807 -1.03548107124723 0.0
-0.404229136435543 -1.0070310013508 0.0
-0.287784411964846 -0.0397957346611695 1.0
-0.0395201915728942 -0.0045068856074113 1.0
0.00161506831413982 0.167879797093616 1.0
0.148940664228943 0.00178756394267521 1.0
0.152260616587067 0.0804453212579361 1.0
0.274285381989209 0.249788264456903 1.0
0.307483193012641 0.271348716044172 1.0
0.454160113460011 0.384770520503727 1.0
0.379940927053089 0.438218168794694 1.0
0.40853186291228 0.669264173502633 1.0
0.493392738672171 0.827651826357994 1.0
0.384662408567469 0.843917358152303 1.0
0.431939902345132 0.898480322311085 1.0
0.303119265801782 1.07950822827985 1.0
0.453821643554384 1.29121053497301 1.0
0.512006316742393 1.35325835138112 1.0
0.398108836984787 1.49361189710571 1.0
0.481133255015837 1.63105047694138 1.0
0.140833594344003 1.80653126517708 1.0
0.324196318630291 1.7885914603562 1.0
0.231321970943864 1.84994616912983 1.0
0.0708573625053293 1.92294814673344 1.0
-0.174481841747028 2.00437408417365 1.0
-0.0839265469398911 2.04880810102944 1.0
-0.0538597554466167 2.01941252071424 1.0
-0.206109918896302 -0.118792439287932 2.0
-0.263796617806589 -0.0440271972638062 2.0
-0.180008822717377 -0.0191154820674229 2.0
-0.252714369128004 0.154753038875566 2.0
-0.146360861592727 0.04615033225229 2.0
-0.140314508794594 0.0136643392681508 2.0
-0.202985668285536 -1.81370036382855e-05 2.0
-0.221307773970668 0.0614601993754487 2.0
-0.0936331357267495 -0.0818846571689675 2.0
-0.0865610429787562 0.115324986870352 2.0
-0.162132944300565 0.198460036899865 2.0
-0.0292526426406232 0.0361984812934024 2.0
-0.00777980299574255 0.208245403890785 2.0
0.0804654347598988 0.0601761889381183 2.0
0.0519115942578465 0.155189807850933 2.0
-0.146499202132317 0.183015325353839 2.0
-0.0513686700205802 0.276312209647027 2.0
-0.152413788853327 0.352462640824186 2.0
-0.0247465423312647 0.225996413868418 2.0
-0.0233823287262425 0.380664489058929 2.0
0.108643970233493 0.445184079546015 2.0
0.035907044285989 0.310616618515652 2.0
0.150722271938072 0.238994398997356 2.0
0.12187681975218 0.346149495809248 2.0
0.00302710581971399 0.608693573497975 2.0
-0.00799972907374968 0.487757495383411 2.0
0.0877578527524834 0.445435398811882 2.0
0.01828696838057 0.548698958778968 2.0
0.0373057300995732 0.525902753983641 2.0
0.166296966643227 0.612088702569156 2.0
0.206190789080826 0.647154271234857 2.0
0.203345543638937 0.650537106521466 2.0
0.0683189756318581 0.803130091264154 2.0
0.200334091975137 0.82093063745688 2.0
0.132885945658044 0.826945886320008 2.0
0.0258144904661712 0.95708927835021 2.0
0.142307284518693 0.938282963261506 2.0
-0.0187149652218148 1.04389758349345 2.0
0.206325060112431 0.873771805978169 2.0
0.0801308162605947 1.00109140649329 2.0
0.129056682668186 1.19589150774701 2.0
0.234819081658577 0.996525924335586 2.0
0.212423027076362 1.06142827935846 2.0
0.117410084668763 1.2440327581476 2.0
0.0691623858912138 1.28299347539703 2.0
0.163742421947584 1.19492305082507 2.0
0.0246799892211049 1.26869152644438 2.0
0.210711185649519 1.24988421558422 2.0
-0.000887730676151222 1.44156097266813 2.0
0.0309595591197476 1.34938914351217 2.0
0.147936616729493 1.3668889877394 2.0
0.158659985194186 1.35612897011904 2.0
0.276662897515644 1.41363218572035 2.0
0.246472472734942 1.46984880777302 2.0
0.241482590086136 1.46937833328243 2.0
0.0133619094022552 1.73498956654963 2.0
0.073661509612148 1.61085135635185 2.0
0.0896570733437889 1.66791367917458 2.0
-0.0155228509651628 1.68476377591718 2.0
-0.0850596548402829 1.80031740231858 2.0
0.167024258943589 1.72230306340769 2.0
-0.00195782817901113 1.70226191537203 2.0
0.110580698810188 1.76140142738798 2.0
0.11993604776784 1.70398908604059 2.0
0.00749799112985539 1.88195945918958 2.0
0.1033485762785 1.84598835747455 2.0
-0.0987088259993369 1.90437115788428 2.0
-0.0166949295473942 2.00638625713888 2.0
0.119362542051087 1.8999028718953 2.0
0.0357118427422101 1.83917622779721 2.0
-0.145827473007981 1.93809032977626 2.0
-0.0925294079599508 1.91175394218614 2.0
-0.123300973475327 1.97559567313436 2.0
0.0291880662007094 1.78593436176704 2.0
-0.121518220146199 1.93231298007913 2.0
-0.185794389309373 1.95324958299531 2.0
-0.114203865166274 1.9172259501703 2.0
-0.148163235742273 1.99713853841754 2.0
-0.227345253647866 2.09377659211818 2.0
-0.114437347566115 2.00143396987072 2.0
};

\end{axis}

\end{tikzpicture}
\centering
}

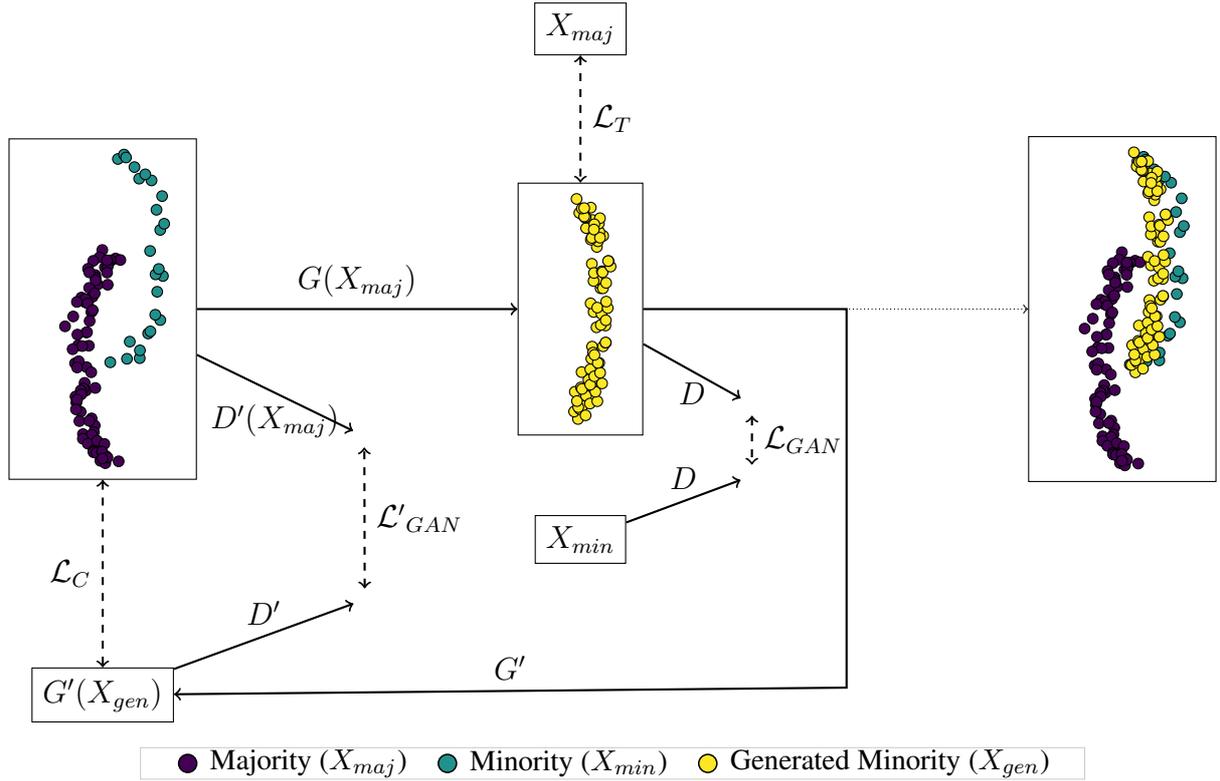
\begin{figure*}
\centering
\begin{tikzpicture}
	\node[rectangle, draw] (initial_maj) {\usebox\initialmajbox};
	
	\node[rectangle, draw, right=10em of initial_maj] (generated) {\usebox\generatedbox};
	\node[rectangle, draw, right=12em of generated] (output) {\usebox\outputbox};
	\node[rectangle, draw, above=4em of generated] (compare_maj) {$X_{maj}$};
	\node[rectangle, draw, below=2.5em of generated] (real) {$X_{min}$};
	\node[rectangle, draw, below=5.9em of initial_maj] (returned_maj) {$G'(X_{gen})$};
	
	\node[right=\diagramlgan of generated, yshift=-3em] (dg) {};
	\node[right=3.57em of real, yshift=2em] (dr) {};
	\node[right=\diagramlgantag of initial_maj, yshift=-4em] (d2r) {};
	\node[right=5.6em of returned_maj, yshift=3em] (d2g) {};
	
	\draw[->, thick] (initial_maj) -- node[above] {$G(X_{maj})$} (generated);
	\draw[->, densely dotted] (generated) -- (output);
	\draw[->, thick] (generated) -- ++(3.5, 0) -- ++(0, \diagramninetydegree) -- node[above] {$G'$} (returned_maj);
	
	\draw[->, thick] (generated) -- node[below] {$D$} (dg);
	\draw[->, thick] (real) -- node[above] {$D$} (dr);
	\draw[->, thick] (initial_maj) -- node[below] {$D'(X_{maj})$} (d2r);
	\draw[->, thick] (returned_maj) -- node[above] {$D'$} (d2g);

	\draw[<->, dashed, thick] (initial_maj) -- node[left] {$\mathcal{L}_C$} (returned_maj);
	\draw[<->, dashed, thick] (compare_maj) -- node[right] {$\mathcal{L}_T$} (generated);
	\draw[<->, dashed, thick] (dg) -- node[right] {$\mathcal{L}_{GAN}$} (dr);
	\draw[<->, dashed, thick] (d2g) -- node[right] {$\mathcal{L'}_{GAN}$} (d2r);


		





\end{tikzpicture}
\vskip 0.1in

\definecolor{majcol}{RGB}{68,1,84}
    \definecolor{mincol}{RGB}{33,145,140}
    \definecolor{gencol}{RGB}{253,231,37}
    \centering
\tikzset{
        leg/.style={
            circle, scale=0.6, line width=0.2mm, draw=black
        },
        }
\begin{tikzpicture}
\begin{axis}[colormap/viridis, ticks=none, width = \legendsizelarge, height = 2cm, xmin=0, xmax=20, ymin=0, ymax=5, draw=white!80!black]
    \node[leg, fill=majcol] at (1,2.5) (blah) {};
    \node[font=\small, right=0.05em of blah] (maj_tag) {Majority ($X_{maj}$)};
    \node[leg, fill=mincol, right=0.6em of maj_tag] (blah2)  {};
    \node[font=\small, right=0.05em of blah2] (min_tag) {Minority ($X_{min}$)};
    \node[leg, fill=gencol, right=0.6em of min_tag] (blah3) {};
    \node[font=\small, right=0.05em of blah3] (gen_tag) {Generated Minority ($X_{gen}$)};
\end{axis}
\end{tikzpicture}

\caption{Training the tabular translation GAN involves imposing additional objectives $\mathcal{L}_T$ (Eq. \ref{eq:lt}), $\mathcal{L}_C$ (Eq. \ref{eq:lc}), and $\mathcal{L}_I$ (Eq. \ref{eq:li}) (in addition to the standard GAN loss (Eq. \ref{eq:lgan})) that help constrain the position of the generated samples with respect to the majority samples. The generated samples are then included in the oversampled dataset. The identity loss $\mathcal{L}_I$ is omitted for the sake of clarity of the diagram.}
\label{fig:architecture}
\end{figure*}

\section{Tabular Translation GAN (TTGAN)}
\subsection{Formulation}
Given a binary imbalanced classification dataset 
$$\mathcal{D} = \{(x_i, y_i) \}_{i=1}^N \text{ where } x_i \in \mathbb{R}^d, y_i \in \{0,1 \},$$ 
we wish to generate new samples to include in an augmented dataset $\mathcal{D'}$. We interchangeably write 
$$\mathcal{D} = X_{maj} \cup X_{min}, $$ where 
$$X_{maj} = \{x_i | y_i=0 \} \text{ and } X_{min} = \{x_i | y_i=1 \},$$ 
are the majority class and minority class respectively.

We refer to the set of synthetic samples we generate as $X_{gen}$ and select a subset $X_{selected}$ of those to add, such that $\mathcal{D'}$ is given by
$$\mathcal{D'} = \mathcal{D} \cup X_{selected},$$ 
for $X_{selected} \subseteq X_{gen}$.

We then fit a classifier $f$ (which may in principle be of any type) on $\mathcal{D'}$ and use it to classify test samples.

\subsection{Generative Adversarial Networks (GAN)}
Goodfellow et al. \yrcite{goodfellow2014generative} introduced a breakthrough generative model to sample from an arbitrary distribution. GANs learn to generate new samples from a desired distribution using two neural networks, the generator $G$ and discriminator $D$, that play a min-max game where the generator is tasked with generating synthetic samples that fool the discriminator, which in turn is tasked with distinguishing between real and generated samples. 

This minimax loss can be formulated as

\begin{equation} \label{eq:lgan}
\mathcal{L}_{GAN} = E_{x}[\log{(D(x))}] + E_{z}[\log{(1-D(G(z)}],
\end{equation}

where $E_x$ is the expected value over real samples $x$, and $E_z$ is the expected value over input noise to the generator $z$ which is sampled from a latent noise space $z \sim Z$ where $Z$ is some prior parametric distribution.

We note that the portion of the loss that is relevant to the generator is
\begin{equation} \label{eq:lg}
\mathcal{L}_G = E_{z}[\log{(1-D(G(z))}]
,
\end{equation}
which the generator tries to minimize and the discriminator tries to maximize.

We denote the portion that is not directly affected by the generator as
\begin{equation*} \label{eq:ld}
\mathcal{L}_D = E_{x}[\log{(D(x))}]
,
\end{equation*}
which the discriminator tries to maximize.

While effective for sampling from the distribution, the standard GAN model provides no mechanism for controlling the location of a generated output sample given a particular input sample. In particular, this means that it is not designed to generate samples for the purpose of  improving downstream imbalanced classification performance. 

The introduction of the regularizing loss functions described below allows us to achieve this goal by performing translation from the majority class. The first such loss directly links between the output sample and the input sample. The additional losses, motivated by CycleGAN \cite{zhu2017unpaired}, link between the output sample and the input sample by way of an additional GAN mapping from the minority to the majority.

\subsection{Direct Translation Loss}

\begin{figure*}
\centering
\input{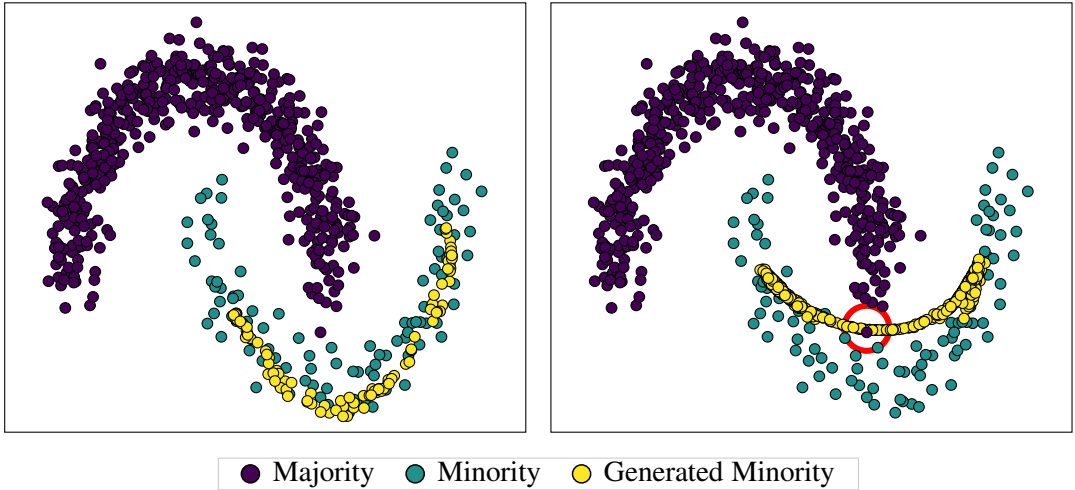}
\vskip 0.1in
\caption{Samples are drawn from a two-moons toy dataset provided by scikit-learn \cite{pedregosa2011scikit}. On the left, the samples are generated by a standard ("vanilla") GAN which is trained to model the minority class distribution. On the right, the tabular translation GAN with $\lambda_T=0.1, \lambda_C=0, \lambda_I=0$ demonstrates the ability of the translation loss to guide the model to generate samples that are close to the class boundary. In the region highlighted by the red circle, some of the generated samples fall within the boundary of the majority class.}
\label{fig:twomoons}
\end{figure*}

For vector $z$, we define the translation loss as

\begin{equation} \label{eq:lt}
\mathcal{L}_T(z) = \lVert z - G(z) \rVert_1
.
\end{equation}

In order to guide the GAN to effectively translate input points to a close synthetic minority sample, we propose the following modifications to the base GAN architecture:

\begin{itemize}
	\item Feeding real majority class samples as inputs to the generator during the training process: $z \sim X_{maj}$ rather than $z \sim Z$. The input $z$ to the generator is no longer a latent random noise vector, but rather an input feature-space sample we wish to translate.
	\item Adding the direct $L_1$ regularization term $\lambda_T \mathcal{L}_T(z)$ (Eq. \ref{eq:lt}) to the generator's loss objective.
\end{itemize}

Intuitively, the addition of the translation loss guides the model to balance its primary objective of modeling the minority distribution with the objective of minimizing the distance between the output translated sample and the input sample. The magnitude of the hyperparameter coefficient $\lambda_T$ modulates this balance.

The effects of introducing these changes can be immediately seen when visualized on synthetic datasets (see \cref{fig:twomoons}), and when constrained with resources and dataset size the direct translation loss suffices.
However, to further leverage the majority samples (beyond using them as inputs for the generator training), we complement the direct translation loss with losses adopted from the CycleGAN \cite{zhu2017unpaired} model.

\subsection{CycleGAN}
Typically used for translating between unpaired domains of images, the CycleGAN \cite{zhu2017unpaired} model introduces an additional pair of networks $G', D'$ responsible for sampling from the majority class distribution. The following losses are added that govern how these networks interact with the networks $G, D$ that we use for sampling from the minority:

\textbf{Cycle-consistency loss -} Encourages the two generators to be roughly inverse mappings of each other:
\begin{equation} \label{eq:lc}
\begin{split}
\mathcal{L}_C(z) =  & \lVert G(G'(x_{min}))-x_{min} \rVert_1  \\ 
+ &  \lVert G'(G(x_{maj}))-x_{maj}\rVert_1 .
\end{split}
\end{equation}
(where $z$ takes on values of $x_{min}$ or $x_{maj}$).

\textbf{Identity loss -} Encourages the generators to be roughly the identity mapping on inputs that are sampled from the output distribution:
\begin{equation} \label{eq:li}
	\begin{split}
\mathcal{L}_I(z)= & \lVert G(x_{min})-x_{min}\rVert_1 \\ 
+ & \lVert G'(x_{maj})-x_{maj}\rVert_1 .
\end{split}
\end{equation}

Intuitively, these losses encourage the translation mechanism to possess the following desirable properties: that by translating back and forth we should arrive (close to) back where we started, and secondly that no translation is needed when already in the desired output domain.

\begin{table*}[hbt!]
	\caption{Notable differences between TTGAN and baseline vanilla GAN}
	\vskip 0.15in
	\centering
	\begin{tabular}{lll}
		\toprule
		\cmidrule(r){1-3}
	&	Vanilla GAN   & TTGAN \\
		\midrule
		Networks & $G, D$ & $G, D, G', D'$ \\
		Input training dist. & $z \sim Z$ & $z \sim X_{maj} $ \\
		Generator loss func. & $\mathcal{L}_G \text{ (Eq. } \ref{eq:lg})$ & $\mathcal{L}_G + \lambda_T \mathcal{L}_T(z) + \lambda_C \mathcal{L}_C(z) + \lambda_I \mathcal{L}_I(z)$  \\
		Synthesized samples & $X_{gen} = G(Z) $ & $X_{gen} = G(X_{maj}) $ \\
		\bottomrule
	\end{tabular}
	\label{tab:outline-differences}
	\vskip -0.1in
\end{table*}

\subsection{Tabular translation GAN}
The full tabular translation GAN consists of generator $G$ and discriminator $D$ which are both fully-connected neural nets. The aforementioned losses are added to the standard GAN loss objectives, such that the full loss function is given by


\begin{equation*}
	\begin{split}
 \mathcal{L'}_G = E_{z}[ & \log{(1-D(G(z))} \\
  + & \lambda_T \mathcal{L}_T(z) + \lambda_C \mathcal{L}_C(z) + \lambda_I \mathcal{L}_I(z)],
	\end{split}
\end{equation*}

and the complete GAN loss is 
\begin{equation} \label{eq:lttgan}
\mathcal{L}_{TTGAN} = \mathcal{L}_D + \mathcal{L'}_G.
\end{equation}

The model modifications thus far is described in \cref{tab:outline-differences}. \cref{fig:architecture} shows the architectural scheme of the training process.

\subsection{Implementation details}

In the experiments we perform, the generator is a 4 layer-deep fully-connected network with hidden sizes of 64, 128, and 256 with SELU \cite{klambauer2017self} nonlinearities. The discriminator is a 3 layer-deep fully-connected network with hidden sizes of 128 and 64 with SELU nonlinearities. Training follows the standard training procedure for GANs, with iterated steps of real and synthetic points, and we use the Adam optimization method with a constant learning rate of $10^{-4}$.

\subsubsection{Synthetic samples selection criterion}

Having trained the tabular translation GAN, we are able to generate synthetic minority samples by applying the generator on the majority class:

$$X_{gen} = G(X_{maj}) .$$

Although in principle we could augment the dataset with the entirety of $X_{gen}$, under certain circumstances this set may contain samples that are too far from the class boundary (in either direction) and thus detrimental. In particular, the additional losses imposed by TTGAN encourage the model to generate samples closer to the class boundary, which have a higher chance to drift over into the majority (as shown in the synthetic dataset of \cref{fig:twomoons}). Our experimentation shows that a selection criterion that filters out samples, described below, yields good results.

We first fit a baseline classifier $f_{b}$ on the input dataset $\mathcal{D}$, such that $f_{b}$ is of the same model type and architecture as the desired final classifier. We use $f_{b}$ to score the likelihood of each of the samples of $X_{gen}$ belonging to the minority class, and we sort the elements of $X_{gen}$ according to this score in descending order. We then apply a cut-off limit $p_{max}$ to filter out samples which have too high of a score (implying they are far from the boundary and thus less useful). Then we  further restrict the number of selected samples to be a desired multiple $s$ of the size of the minority class, specifically:

$$ X_{selected} = sorted(\{ x \in X_{gen} | f_{b}(x) \le p_{max}\}) [:s|X_{min}|].$$
We consider both the cut-off probability $p_{max}$ and the number of samples to retain $s$ (as a multiple of the size of the minority class) to be hyperparameters of the model (see \cref{sec:experiments} for details on tuning).

The resulting augmented, resampled dataset is then given by
$$ \mathcal{D'} = X_{maj} \cup X_{min} \cup X_{selected}.$$

The entire proposed procedure is detailed in \cref{alg:alg}.

\begin{algorithm}[hbt!]
\caption{Classification with Tabular Translation GAN}\label{alg:alg}
\begin{algorithmic}[1]
\renewcommand{\algorithmicrequire}{\textbf{Input:}}
\renewcommand{\algorithmicensure}{\textbf{Output:}}
\REQUIRE{Binary classification dataset $\mathcal{D} = X_{maj} \cup X_{min} $}
\ENSURE{Classifier $f$}
\STATE Fit baseline classifier $f_b$\;
\STATE Train tabular translation GAN using $\mathcal{L}_{TTGAN}$ yielding generator $G$\;
\STATE Translate $X_{gen} \gets G(X_{maj})$\;
\STATE Filter synthetic points samples \\
$sort(\{ x \in X_{gen} | f_{b}(x) \le p_{max}\}) [:s|X_{min}|]$\ \\
and assign the result to $X_{selected}$\;
\STATE Augment (oversample) dataset $\mathcal{D'} \gets \mathcal{D} \cup X_{selected}$\;
\STATE Fit final classifier $f$ on augmented dataset $\mathcal{D'}$\;
\end{algorithmic}
\end{algorithm}


\begin{table*}[hbt!]
    \fontsize{10pt}{12pt}\selectfont
	\caption{Classification performance on KEEL datasets}
	\centering
	\vskip 0.15in
	\begin{tabular}{lllllllll}
		\toprule
		\multicolumn{3}{c}{Dataset characteristics}  &     \multicolumn{3}{c}{mAP (rank)}             \\
		\cmidrule(r){1-3}
		Dataset   & N & IR & RW & ROS  & SMOTE & B-SMOTE & SUGAR & TTGAN \\
		\midrule
		abalone9-18 & 4174 & 129  & 0.8364 (3) & 0.7936 (5) & 0.8025 (4) & \textbf{0.8744} (1) & 0.7415 (6) & 0.8304 (2) \\
		abalone19   & 731 & 16 & 0.0081 (6) & 0.0217 (2) & 0.0212 (3) & 0.0124 (4) & 0.0121 (5) & \textbf{0.0363} (1)   \\
		glass-0-1-6\_vs\_2 & 192 & 10 & 0.1958 (6) & \textbf{0.3923} (1) & 0.3250 (4) & 0.3583 (2) & 0.3310 (3) & 0.2386 (5) \\
		glass2 & 214 & 11 & \textbf{0.1942} (1) & 0.1517 (5) & 0.1557 (3) & 0.1520 (4) & 0.1008 (6) & 0.1909 (2) \\
		glass4 & 214 & 15 & 0.3417 (3.5) & 0.3000 (6) & 0.3595 (3) & \textbf{0.4444} (1) & 0.3417 (3.5) & 0.3833 (2) \\
		page-blocks-1-3\_vs\_4 & 472 & 16 & 0.7958 (3) & 0.7579 (5) & 0.7579 (5) & 0.7579 (5) & 0.8076 (2) & \textbf{0.8972} (1) \\
		yeast-0-5-6-7-9\_vs\_4 & 528 & 9 & 0.7284 (2) & 0.5421 (5) & 0.6245 (4) & 0.6911 (3) & 0.5380 (6) & \textbf{0.7502} (1) \\
		yeast-1\_vs\_7 & 459 & 30 & 0.1980 (6) & \textbf{0.4043} (1) & 0.3076 (3) & 0.3709 (2) & 0.2800 (5) & 0.2926 (4) \\
		yeast-1-2-8-9\_vs\_7 & 947 & 22 & 0.2355 (5) &  0.2499 (4) & 0.2647 (2) & \textbf{0.2677} (1) & 0.2206 (6) & 0.2546 (3) \\
		yeast-1-4-5-8\_vs\_7 & 693 & 14 & 0.0591 (6) & 0.1084 (4) & 0.1589 (2) & \textbf{0.2416} (1) & 0.0793 (5) & 0.1294 (3) \\
		yeast-2\_vs\_4 & 514 & 9 & \textbf{0.8563} (1) & 0.8467 (2) & 0.7892 (4) & 0.7540 (5) & 0.7080 (6) & 0.8162 (3) \\
		yeast-2\_vs\_8 & 482 & 23 & 0.5294 (6) & 0.5481 (4) & 0.5450 (5) & 0.5482 (3) & 0.5518 (2) & \textbf{0.5573} (1) \\
		yeast4 & 1484 & 28 & 0.4550 (4) & 0.4525 (5) & 0.4310 (6) & 0.5738 (2) & 0.4989 (3) & \textbf{0.5770} (1) \\
		yeast5 & 1484 & 32 & 0.7326 (2) & 0.6620 (6) & 0.6865 (5) & 0.6888 (4) & \textbf{0.7419} (1) & 0.7170 (3) \\
		yeast6 & 1484 & 41 & 0.5986 (2) & 0.4959 (4) & 0.5095 (3) & 0.3818 (6) & 0.4504 (5) & \textbf{0.6278} (1) \\
		\midrule
		Mean &  &  & 0.451 & 0.448 & 0.449 & 0.474 & 0.427 & \textbf{0.487} \\
		Mean rank &  &  & 3.767 & 3.933 & 3.733 & 2.933 & 4.3 & \textbf{2.2} \\
		Median rank &  &  & 3.5 & 4 & 4 & 3 & 5 & 2 \\
		\# of datasets best &  &  & 2 & 2 & 0 & 4 & 1 & \textbf{6} \\
		\# of datasets worst &  &  & 5 & 2 & 1 & 1 & 5 & \textbf{0} \\
		\bottomrule
	\end{tabular}
	\label{tab:keel-results}
	\vskip -0.1in
\end{table*}

\section{Experiments}
\label{sec:experiments}

We evaluate the classification performance of the proposed model under several different regimes of classifier model type, dataset, and imbalance ratio. 

The alternative approaches that we compare to are: Re-weighting (RW) - the baseline approach where the chosen classifier is trained with re-weighting applied on the loss function to maximize performance on a balanced testing criterion, Random Oversampling (ROS) where the minority class is randomly oversampled so as to match the size of the majority class, SMOTE \cite{chawla2002smote}, Borderline SMOTE (B-SMOTE) \cite{han2005borderline}, and SUGAR \cite{lindenbaum2018geometry}.

For each experiment, we record values of the mean average precision (mAP) metric. In the result tables, we also note characteristics of the datasets ($N$ = \# of samples, IR = imbalance ratio). For the tabular translation GAN model, choices for the loss coefficient hyperparameters and selection hyperparameters are chosen via a hyperparameter search using Optuna \cite{akiba2019optuna} on a validation set. Hyperparameter specifications are detailed in \cref{appendix}.

In all experiments, pre-processing consisted of normalizing the features and handling categorical features. In the first and second set of experiments, categorical features were one-hot encoded, while in the third set the categorical features were target-encoded using the Category Encoders \cite{mcginnis2018category} package. In addition, in the third set missing feature values were imputed with the mode of the feature, and the Yeo-Johnson transformation \cite{yeo2000new} was applied to features that were determined to follow a power law distribution - specifically,  if over 90\% of the data falls in less than 20\% of the value range of the feature. 

For the first set of experiments, we test the performance of a linear SVM classifier on the KEEL \cite{alcala2011keel} collection of datasets. The scikit-learn \cite{pedregosa2011scikit} implementation is used and loss weights are balanced accordingly to compensate for imbalance. Examining the results in \cref{tab:keel-results}, we find that the proposed model outperforms all others in terms of mAP and median rank, performs the best on the highest number of datasets and is not the worst-performing approach on any dataset.

\begin{table*}[hbt!]
    \fontsize{11pt}{13pt}\selectfont
	\caption{Classification performance on CelebA dataset}
	\vskip 0.15in
	\centering
	\begin{tabular}{lllllllll}
		\toprule
		\multicolumn{3}{c}{Dataset characteristics}  &     \multicolumn{3}{c}{mAP}             \\
		\cmidrule(r){1-3}
		Dataset   & N & IR & RW & ROS & SMOTE & B-SMOTE & SUGAR & TTGAN \\
		\midrule
		CelebA & 203k & 40  & 0.4431 & 0.3918 & 0.3885 & 0.3773 & 0.4062 & \textbf{0.4444} \\
		CelebA (extreme imba.)   & 199k & 200 & 0.1435 & 0.1243 & 0.0944 & 0.0981 & 0.1254 & \textbf{0.1445}   \\
		\bottomrule
	\end{tabular}
	\label{tab:celeba-results}
	\vskip -0.1in
\end{table*}

\begin{table*}
    \fontsize{11pt}{13pt}\selectfont
	\caption{Classification performance on Playtika datasets}
	\vskip 0.15in
	\centering
	\begin{tabular}{lllllllll}
		\toprule
		\multicolumn{3}{c}{Dataset characteristics}  &     \multicolumn{2}{c}{mAP} & \multicolumn{2}{c}{AUC-ROC} & \multicolumn{2}{c}{Precision @ 0.4 Recall}          \\
		\cmidrule(r){1-3}
		\cmidrule(r){4-5}
		\cmidrule(r){6-7}
		\cmidrule(r){8-9}
		Dataset   & N & IR  & Baseline & TTGAN & Baseline & TTGAN & Baseline & TTGAN\\
		\midrule
		Churn & 6.3M & 23  & 0.255 & \textbf{0.266} & 0.852 & \textbf{0.856} & 0.276 & \textbf{0.282} \\
		Task 1 Return   & 45.3M & 200  & 0.12 & 0.12 & 0.928 &  0.928 & 0.132 & 0.132 \\
		Task 1 Pay & 45.3M & 200  & 0.215 & \textbf{0.219} & 0.946 & 0.946 & 0.239 & 0.239 \\
		Task 2 Return & 60M & 220  & \textbf{0.093} & 0.092 & 0.918 & 0.918 & 0.104 & \textbf{0.105} \\
		Task 2 Pay & 60M & 220  & 0.127 & \textbf{0.132} & 0.946 & \textbf{0.947} & 0.105 & \textbf{0.113} \\
		\bottomrule
	\end{tabular}
	\label{tab:exp3-results}
	\vskip 0.1in
\end{table*}

The second set of experiments involves applying the state-of-the-art Catboost \cite{catboost} boosted trees model on a version of the large CelebA dataset where the features are annotated attributes rather than pixel values, and the target is another attribute \cite{liu2015faceattributes}. We also use an artificially extreme-imbalanced version of this dataset derived by withholding minority samples. Catboost is trained with the F1 loss function, a learning rate of 0.2, for 50 iterations with depth 3 trees. The results are recorded in \cref{tab:celeba-results}.
Here we see a clear advantage of the robustness of the proposed approach when compared to the alternatives: while the other approaches suffer degradation compared to the baseline approach, the tabular translation GAN instead provides a marginal performance improvement over the strong baseline model.

In the third set of experiments, we tested our method on datasets collected by Playtika \footnote{https://www.playtika.com/}. Playtika collected and preprocessed datasets over the course of several months concerning customer behavior. The purpose of these datasets is to perform several classification tasks: predicting customer churn (turnover), returning within a specified time interval (Return) and probability to pay within a specified time interval (Pay). Each of these datasets is very large, ranging from 6.3 million samples to 60 million samples, and they each exhibit imbalance ranging from an imbalance ratio of 23 to an extreme imbalance ratio of 220. Improving upon the Catboost-based baseline on these challenging tasks directly affects the ability of the organization to better understand its customers and to take appropriate steps to cater to them. We measure performance on the metrics of AUC-ROC, mAP, and Precision @ 0.4 recall. In the results shown in  \cref{tab:exp3-results}, we see a noticeable improvement on the primary dataset of focus, customer churn. The other datasets exhibit performance that either remains the same or more modestly improves; these datasets are large with varying characteristics and exhibit extreme imbalance. We also note that the mechanism of synthetic sample selection was slightly modified during these experiments: instead of $p_{max}$ representing an upper bound, samples were selected to be as close as possible to $p_{max}$ on either side of it; this improved performance but the interpretation of $p_{max}$ as a parameter that constrains the sample threshold remains unchanged.

Lastly, we also use the UMAP \cite{mcinnes2018umap-software} dimensionality reduction technique to visualize samples from the yeast4 \cite{alcala2011keel} dataset to better understand classification improvement. We compare an identical number of synthetic samples generated by a vanilla GAN and TTGAN respectively. This visualization, shown in \cref{fig:vis}, demonstrates that TTGAN leads the synthesized points to be closer to the class boundary than with a vanilla GAN. A results comparison is given in \cref{tab:vis}.

\begin{figure*}
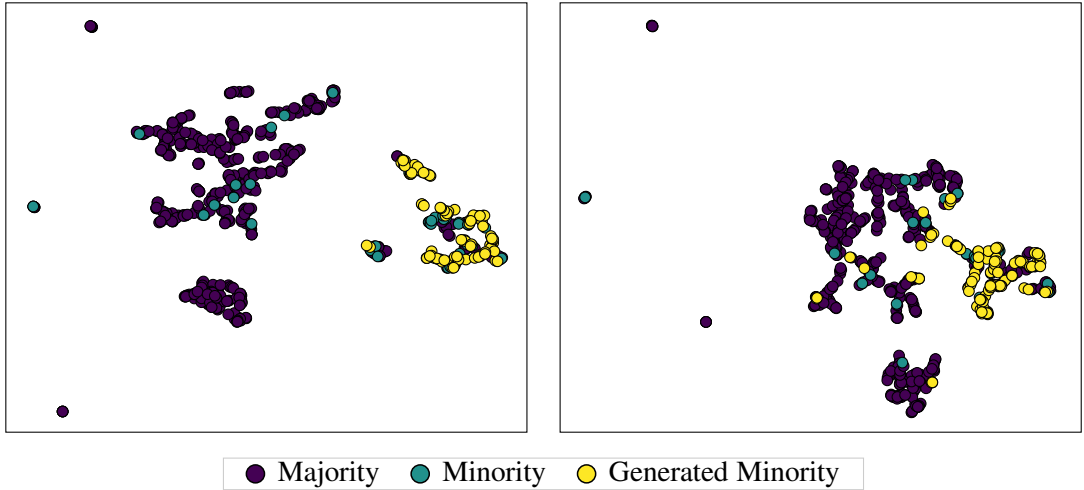

	\centering
    \input{plots/plot_vanilla_umap_yeast4.tex}
    \hspace{3pt}
    \input{plots/plot_ttgan_umap_yeast4.tex}
    \vskip 0.1in
    \definecolor{majcol}{RGB}{68,1,84}
    \definecolor{mincol}{RGB}{33,145,140}
    \definecolor{gencol}{RGB}{253,231,37}
    \centering
\tikzset{
        leg/.style={
            circle, scale=0.6, line width=0.2mm, draw=black
        },
        }
\begin{tikzpicture}
\begin{axis}[colormap/viridis, ticks=none, width = \legendsize, height = 2cm, xmin=0, xmax=20, ymin=0, ymax=5, draw=white!80!black]
    \node[leg, fill=majcol] at (1,2.5) (blah) {};
    \node[font=\small, right=0.05em of blah] (maj_tag) {Majority};
    \node[leg, fill=mincol, right=0.6em of maj_tag] (blah2)  {};
    \node[font=\small, right=0.05em of blah2] (min_tag) {Minority};
    \node[leg, fill=gencol, right=0.6em of min_tag] (blah3) {};
    \node[font=\small, right=0.05em of blah3] (gen_tag) {Generated Minority};
\end{axis}
\end{tikzpicture}
\vskip 0.1in
	\caption{A 2-dimensional visualization using UMAP \cite{mcinnes2018umap-software} of samples from the yeast4 dataset \cite{alcala2011keel}. On the left, synthetic samples are generated by a vanilla GAN; on the right, synthetic samples are generated by TTGAN. TTGAN generates samples from more diverse regions, closer to the class boundary.}
	\label{fig:vis}
\end{figure*}

\subsection{Ablation analysis}

To further analyze the performance of the proposed \cref{alg:alg}, we conduct an ablation test by iteratively removing components of the model: comparing TTGAN to TTGAN without the cyclic and identity losses, and comparing that to a "vanilla" GAN model.
We conduct these tests on the page-blocks-1-3\_vs\_4 dataset from \cite{alcala2011keel}. For all experiments besides the baseline, the number of samples generated was 5.5x the number of original minority, as determined by a hyperparameter search on a validation set. The results are shown in \cref{tab:ablation}, and show that removing components of the model degrades performance.

\begin{table}
	\caption{Linear SVM yeast4 comparison}
	\vskip 0.15in
	\centering
	\begin{tabular}{ll}
		\toprule
		Model   & mAP \\
		\midrule
		Vanilla GAN   & 0.5352   \\
		TTGAN & \textbf{0.5770} \\
		\bottomrule
	\end{tabular}
	\label{tab:vis}
	\vskip -0.1in
\end{table}

\begin{table}
	\caption{Linear SVM page-blocks ablation analysis}
	\vskip 0.15in
	\centering
	\begin{tabular}{ll}
		\toprule
		Model   & mAP \\
		\midrule
		Baseline & 0.7958 \\
		Vanilla GAN   & 0.8069   \\
		TTGAN w/o cyclic+identity losses & 0.8357 \\
		TTGAN & \textbf{0.8972} \\
		\bottomrule
	\end{tabular}
	\label{tab:ablation}
	\vskip -0.1in
\end{table}

We also note that all datasets in all experiments yielded hyperparameters such that the translation loss and cyclic loss were never simultaneously zero, despite that being an option in the hyperparameter search. This suggests improvement over a vanilla GAN trained on the minority class.

\section{Conclusion}
We presented an end-to-end method for performing imbalanced classification on tabular data, by introducing a GAN that is capable of translating majority samples to minority samples and then using it to generate useful synthetic samples of the minority class. 
We described the modifications to the GAN framework necessary to achieve the desired effect, as well as the mechanism via which we choose which of the translated samples we use.
The experimental results presented show that this approach improves classification performance on a variety of datasets with different characteristics and with different types of underlying downstream classifiers, and that the results remains relatively robust without suffering from degradation.
\subsection{Future work}
While applied here to binary classification on tabular datasets, the proposed approach could in principle be modified, applied and tested in other settings such as multi-class classification, or on different modalities of data. In addition, alternative generative models (such as Variational Auto Encoders \cite{kingma2013auto, li2020variational}, diffusion models \cite{ho2020denoising} or normalizing flows \cite{kobyzev2020normalizing}) could perhaps be modified with the proposed regularization losses in place of the GAN.

\section{Acknowledgments}
This research was partially supported by the Israel Science Foundation (ISF, 1556/17, 1873/21), Israel Ministry of Science Technology and Space 3-16414, 3-14481, and by Consortium (Magnaton), between Playtika and Tel Aviv University and Innovation Authority 69316.

\bibliographystyle{unsrtnat}
\bibliography{arxiv}

\newpage
\appendix
\section{Appendix}
\label{appendix}
The full hyperparameter specification for the first set of experiments is described in \cref{tab:hyperparams}, and the hyperparameters for the third set of experiments is described in \cref{tab:hyperparams-exp3}.

\begin{table}[hbt!]
	\caption{TTGAN KEEL hyperparameters - w / Linear SVM}
	\vskip 0.15in
	\centering
	\begin{tabular}{lllllll}
		\toprule
		Dataset   & Epochs & $\lambda_T$  & $\lambda_C$ & $\lambda_I$ & $s$ & $p_{max}$\\
		\midrule
		abalone9-18 & 1150 & 0.1  & 0 & 0 & 4 & 0.8 \\
		abalone19   & 250  & 0.05  & 10 & 5 & 16 & 0.9  \\
		glass-0-1-6\_vs\_2 & 2500 & 0.05  & 15 & 0 & 5.5 & 0.9 \\
		glass2 & 2500 & 0.1  & 0 & 2.5 & 6.5 & 0.8 \\
		glass4 & 900 & 0  & 15 & 7.5 & 5.5 & 0.8 \\
		page-blocks-1-3\_vs\_4 & 900 & 0.05  & 5 & 5 & 5.5 & 0.7 \\
		yeast-0-5-6-7-9\_vs\_4 & 900 & 0  & 5 & 2.5 & 6.5 & 0.8 \\
		yeast-1\_vs\_7 & 2500 & 0.05  & 0 & 0 & 1.3 & 0.8 \\
		yeast-1-2-8-9\_vs\_7 & 1150 & 0.05  & 10 & 5 & 4 & 1 \\
		yeast-1-4-5-8\_vs\_7 & 900 & 0.1  & 15 & 0 & 1.65 & 0.7 \\
		yeast-2\_vs\_4 & 500 & 0.1  & 15 & 0 & 1.8 & 0.6 \\
		yeast-2\_vs\_8 & 1300 & 0.05  & 10 & 10 & 4 & 0.6 \\
		yeast4 & 1000 & 0.05  & 10 & 0 & 4 & 1 \\
		yeast5 & 1450 & 0.05  & 0 & 0 & 4 & 0.6 \\
		yeast6 & 2500 & 0.15  & 0 & 5 & 7.5 & 0.6 \\
		\bottomrule
	\end{tabular}
	\label{tab:hyperparams}
	\vskip -0.1in
\end{table}

\begin{table}[hbt!]
	\caption{TTGAN Experiment 3 hyperparameters - w / Catboost}
	\vskip 0.15in
	\centering
	\begin{tabular}{lllllll}
		\toprule
		Dataset   & Epochs & $\lambda_T$  & $\lambda_C$ & $\lambda_I$ & $s$ & $p_{max}$\\
		\midrule
		Churn & 700 & 0.2  & 20 & 12 & 0.33 & 0 \\
		Task 1 Return   & 500  & 0  & 4 & 6 & 0.215 & 0  \\
		Task 1 Pay & 700 & 0.05  & 16 & 0 & 0.24 & 0.85 \\
		Task 2 Return  & 700 & 0.05 & 10 & 6 & 0.25 & 0.5\\
		Task 2 Pay & 700 & 0.25  & 10 & 3 & 0.75 & 0.5 \\
		\bottomrule
	\end{tabular}
	\label{tab:hyperparams-exp3}
	\vskip -0.1in
\end{table}

\end{document}